%% file: main.tex
\documentclass[10pt,twocolumn,letterpaper]{article}

\usepackage[pagenumbers]{cvpr}  

\usepackage{capt-of}
\usepackage{xspace}
\usepackage{xcolor}
\usepackage{enumitem}
\usepackage{amsmath,amsthm,amssymb,amsfonts,dsfont,pifont,bm,bbm,mathrsfs,mathtools,nicefrac,extarrows,relsize}
\usepackage{algorithm,algpseudocode,listings}
\usepackage{booktabs,multirow,adjustbox,diagbox,threeparttable,tabularray,setspace}

\definecolor{cvprblue}{rgb}{0.21,0.49,0.74}
\usepackage[pagebackref,breaklinks,colorlinks,citecolor=cvprblue,bookmarks=false]{hyperref}
\usepackage{wrapfig}
\usepackage[capitalize]{cleveref}  

\crefname{section}{Sec.}{Secs.}
\Crefname{section}{Section}{Sections}
\crefname{appendix}{App.}{Apps.}
\Crefname{appendix}{Appendix}{Appendices}
\crefname{table}{Tab.}{Tabs.}
\Crefname{table}{Table}{Tables}
\crefname{figure}{Fig.}{Figs.}
\Crefname{figure}{Figure}{Figures}
\crefname{equation}{Eq.}{Eqs.}
\Crefname{equation}{Equation}{Equations}
\crefname{theorem}{Thm.}{Thms.}
\Crefname{theorem}{Theorem}{Theorems}
\crefname{lemma}{Lem.}{Lems.}
\Crefname{lemma}{Lemma}{Lemmas}
\crefname{remark}{Rem.}{Rems.}
\Crefname{remark}{Remark}{Remarks}
\crefname{corollary}{Cor.}{Cors.}
\Crefname{corollary}{Corollary}{Corollaries}
\crefname{algorithm}{Alg.}{Algs.}
\Crefname{algorithm}{Algorithm}{Algorithms}
\definecolor{cellred}{RGB}{213, 123, 101}
\definecolor{cellgreen}{RGB}{0, 205, 0}
\definecolor{cellblue}{RGB}{54, 125, 189}
\definecolor{codegreen}{rgb}{0,0.6,0}
\definecolor{codegray}{rgb}{0.5,0.5,0.5}
\definecolor{codepurple}{rgb}{0.58,0,0.82}
\definecolor{backcolour}{rgb}{1.0,1.0,1.0}
\lstdefinestyle{mystyle}{
    backgroundcolor=\color{backcolour},
    commentstyle=\color{codegreen},
    keywordstyle=\color{magenta},
    numberstyle=\tiny\color{codegray},
    stringstyle=\color{codepurple},
    basicstyle=\ttfamily\scriptsize,
    breakatwhitespace=false,
    breaklines=true,
    captionpos=b,
    keepspaces=true,
    numbers=left,
    numbersep=5pt,
    showspaces=false,
    showstringspaces=false,
    showtabs=false,
    tabsize=2
}
\newcommand{\tocite}[1]{{\color{red} [TO CITE]}}

\def\paperID{2144}
\def\confName{CVPR}
\def\confYear{2025}

\def\etal{{\it{et~al.\xspace}}}

\title{VanGogh: A Unified Multimodal Diffusion-based Framework for Video Colorization}

\author{
  Zixun Fang$^{1}$ \quad Zhiheng Liu$^{2}$ \quad Kai Zhu$^{1}$ \quad Yu Liu$^{4}$ \quad Ka Leong Cheng$^{3}$ \\ Wei Zhai$^{1\dag}$ \quad Yang Cao$^{1}$ \quad Zheng-Jun Zha$^{1}$\\
\\
  $^1$USTC \quad
  $^2$HKU \quad
  $^3$HKUST  \quad
  $^4$Independent Researcher
}

\begin{document}

\twocolumn[{
\renewcommand\twocolumn[1][]{#1}
\maketitle
\begin{center}
    \vspace{-0.8cm}
    \includegraphics[width=0.93\linewidth]{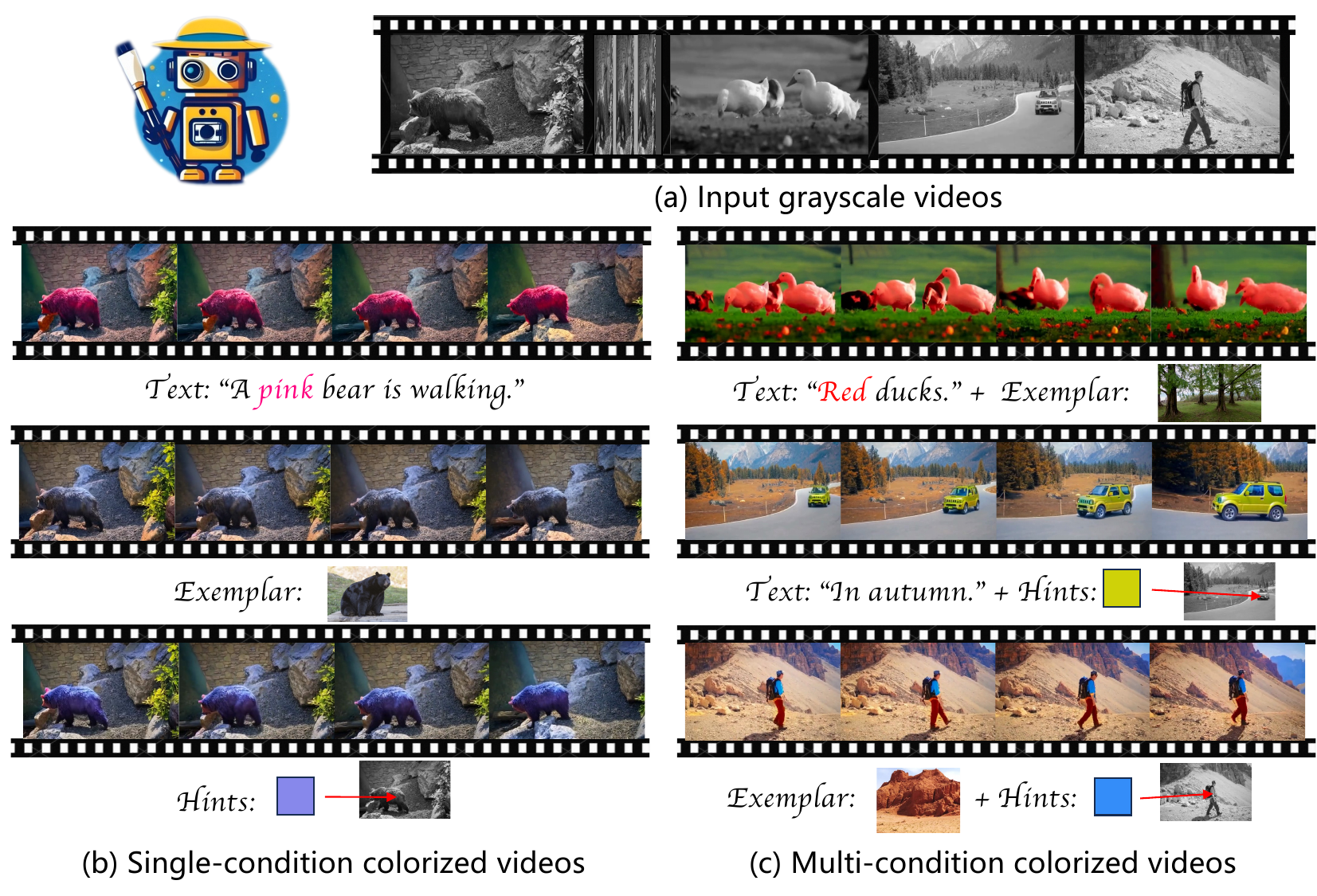}
    \captionsetup{type=figure}
    \vspace{-0.25cm}
    \caption{%
       We present \textbf{VanGogh}, a novel multimodal video colorization method that accepts one or more conditional inputs to generate high-quality, vivid colorization results. Given the grayscale video in (a), our method can take a single condition as input to produce the result shown in (b), or accept multiple conditions for joint control, as demonstrated in (c). Please zoom in for more details.
    }
    \label{fig:teaser}
\end{center}
}]

\input{sections/0_abstract.tex}
\input{sections/1_intro.tex}

\input{sections/2_related.tex}
\input{sections/3_method.tex}

\input{sections/4_exp.tex}

\input{sections/5_conclusion.tex}

\input{sections/6_ref.tex}

\input{sections/7_appendix_1.tex}

\end{document}

%% file: sections/0_abstract.tex
\begin{abstract}
Video colorization aims to transform grayscale videos into vivid color representations while maintaining temporal consistency and structural integrity. 
Existing video colorization methods often suffer from color bleeding and lack comprehensive control, particularly under complex motion or diverse semantic cues. 
To this end, we introduce \textit{VanGogh}, a unified multimodal diffusion-based framework for video colorization. 
VanGogh tackles these challenges using a Dual Qformer to align and fuse features from multiple modalities, complemented by a depth-guided generation process and an optical flow loss, which help reduce color overflow. 
Additionally, a color injection strategy and luma channel replacement are implemented to improve generalization and mitigate flickering artifacts. 
Thanks to this design, users can exercise both global and local control over the generation process, resulting in higher-quality colorized videos. 
Extensive qualitative and quantitative evaluations, and user studies, demonstrate that VanGogh achieves superior temporal consistency and color fidelity. Project page: \url{https://becauseimbatman0.github.io/VanGogh}.

\end{abstract}

%% file: sections/1_intro.tex
\section{Introduction}\label{sec:intro}

Video colorization~\cite{liu2018switchable,zhang2019deep,iizuka2019deepremaster,yang2024bistnet,yang2025colormnet,zhao2023svcnet,liu2023video,li2024towards,zhao2022vcgan,vondrick2018tracking} aims to transform a grayscale video into a colorized version while maintaining temporal consistency across frames and preserving the original structural information. 
This technology enhances the expressive quality and aesthetic appeal of videos, with broad applications in areas such as vintage video restoration and artistic creation.


Despite previous advancements, two significant challenges persist: the lack of comprehensive controllability and color bleeding.
Automatic colorization techniques~\cite{zhao2022vcgan,lei2019fully} generate natural results based on model priors, but they struggle to align with user-provided conditions.
Text~\cite{liu2023video,li2024towards} and exemplar-based~\cite{liu2018switchable,zhang2019deep,iizuka2019deepremaster,vondrick2018tracking,yang2024bistnet,yang2025colormnet} methods offer a global, semantic-level guidance, but they lack user-interactive, fine-grained control capabilities.
Recent studies introduce multi-modal image colorization~\cite{huang2022unicolor,liang2024control,bozic2024versatile} by integrating conditions such as text, exemplars, and hints. This approach allows for accurate color allocation and spatial orientation, fostering greater user interaction and markedly elevating overall quality.
How to effectively integrate multi-modal guidance in video coloring while maintaining the inherent properties of video remains an open question.

Recently, diffusion-based video generative models~\cite{li2023videogen,blattmann2023stable,xing2025dynamicrafter} show significant advancements due to their extensive training on large datasets of text-video pairs. These models possess rich priors and exhibit strong cross-modal capabilities.
In this work, we choose Stable Video Diffusion (SVD)~\cite{blattmann2023stable} as the base model. To better align the text and image conditions, we design a lightweight Dual Qformer structure that extracts features from text and image respectively and fuses them in a shared feature space.
%
Notably, to enable the model to better learn color distribution from reference images, we introduce a color injection strategy that decouples color distribution from structural information, thus enhancing generalization capability. 
For hint conditions, we directly mark hints in grayscale videos and feed them into the UNet, avoiding the complexity of a ControlNet-style~\cite{zhang2023adding} approach.

However, we find that directly applying this method to colorize videos results in significant color bleeding and color overflow. 
%
Furthermore, we observe that color overflow significantly impacts optical flow estimation, thereby degrading video quality.
To address this issue, we design a flow-based perceptual loss that randomly selects two predicted video frames for optical flow estimation~\cite{teed2020raft,xu2022gmflow}, followed by loss calculation against the ground truth to reduce color overflow at object edges. We also incorporate video depth as an aid to enhance color-spatial consistency.
%
Additionally, to mitigate the issue of imprecise reconstruction in certain detail areas of current video VAEs~\cite{zhao2024cv,ge2022long}, which leads to flickering artifacts, we utilize known grayscale videos for luma channel replacement in $Lab$ color space during inference, thereby improving video quality.

%

In summary, our primary contributions are as follows:
1) We propose a novel multimodal video colorization framework that supports not only automatic colorization but also allows for combinations of one or more of the following types of guidance: text, exemplars, and hints, effectively enhancing the flexibility and interactivity of video colorization.
2) We employ a Dual Qformer to integrate images and text, along with a Depth Guider and optical flow loss to alleviate the color overflow effect. 
Additionally, we utilize luma channel replacement to address the inherent flickering artifacts in video VAEs, thereby enhancing the visual quality of the generated results.
3) We design qualitative and quantitative experiments, as well as user studies, to demonstrate the effectiveness and superiority of our method.

%% file: sections/2_related.tex
\section{Related Work}\label{sec:related}
\noindent\textbf{Image colorization.} Image colorization approaches~\cite{larsson2016learning,cheng2015deep,deshpande2015learning,zhang2016colorful,su2020instance,deshpande2017learning,weng2022ct,kang2023ddcolor,cong2024automatic,liang2024control,huang2022unicolor,bozic2024versatile,yun2023icolorit,bai2022semantic,ji2022colorformer,zabari2023diffusing} can be divided into two categories: automatic colorization and conditional colorization. 
Automatic colorization~\cite{cheng2015deep,larsson2016learning,zhang2016colorful,kang2023ddcolor,su2020instance,cong2024automatic,ji2022colorformer,deshpande2015learning,deshpande2017learning,weng2022ct} aims to convert a grayscale image into a natural, vivid color image without user intention. 
Previous researchers~\cite{zhang2016colorful} train a CNN to map from a grayscale image to a color distribution as they treat the problem as multinomial classification. 
Instance-aware colorization~\cite{su2020instance} is achieved by leveraging off-the-shelf models to detect object features, resulting in smooth outputs.
ColorFormer~\cite{ji2022colorformer} proposes a novel framework that utilizes hybrid attention and color memory to extract contextual semantics and diverse color acquisition.
DDColor~\cite{kang2023ddcolor} employs dual decoders: one learns color queries from visual features, while the other provides multi-scale semantic representations.
A more recent work~\cite{cong2024automatic} introduces an imagination module based on powerful diffusion models to synthesize diverse and colorful outcomes.
Although the methods mentioned above can achieve natural and pleasing results, shortcomings, including grayish tones and color bleeding, still exist. 
Additionally, the inability to receive user intervention as guidance limits the application of automatic colorization. 
As a result, many studies explore conditional colorization~\cite{huang2022unicolor,zabari2023diffusing,bai2022semantic,yun2023icolorit,bozic2024versatile,liang2024control,fang2019superpixel,li2019automatic,xu2020stylization,wang2023unsupervised,weng2024cad,chang2023coins}, aiming to generate high-quality color images that align with given conditions such as text, exemplar images, and hints.
Diffusing Colors~\cite{zabari2023diffusing} analyzes the color properties of the VAE latent space and achieves superior control through textual cues for color guidance.
For exemplar-based methods,~\cite{bai2022semantic} builds a more accurate correspondence between a coarse result and the reference image to produce more detailed results with lower computation cost. 
Ke \etal~\cite{ke2023neural} treat the methods in reference-guided
colorization as style transfer works~\cite{gatys2015neural,yoo2019photorealistic,men2022unpaired}.
iColoriT~\cite{yun2023icolorit} utilizes a ViT for hints-interactive colorization and achieves real-time performance.
More recently,~\cite{huang2022unicolor,bozic2024versatile,liang2024control} integrate text, exemplar, and scribble to perform colorization in a multi-modal manner, significantly enhancing the flexibility and interactivity of the colorization process. 
Although the aforementioned methods achieve great success in image colorization, applying them to the video domain can lead to flaws such as temporal inconsistency, error accumulation, and color bleeding.

\noindent\textbf{Video colorization.} Due to the additional time dimension, video colorization~\cite{li2024towards,zhao2023svcnet,yang2025colormnet,yang2024bistnet,lei2019fully,liu2023video,liu2018switchable,zhang2019deep,iizuka2019deepremaster,zhao2022vcgan,vondrick2018tracking} requires not only that the color of each individual frame is visual-appealing, but also that the relationships between frames are considered. 
Lei \etal~\cite{lei2019fully} propose a self-regularized approach to automatic video colorization.
However, it still tends to wash out the colors and fails to respond to user intent.
Text-based video colorization methods utilize human language to guide the coloring process. 
~\cite{li2024towards,liu2023video} employ text-to-image diffusion models to combine text conditions with grayscale videos.
Despite many efforts, there are still issues with color overflow and temporal incoherence. 
Moreover, text conditions cannot achieve fine-grained local control and only provide semantic-level color descriptions such as \textit{"red"} or \textit{"yellow"}.
Similar to text-based methods, exemplar-based video colorization aims to generate colorized videos that align with the target exemplars. 
ColorMNet~\cite{yang2025colormnet} develops a memory-based feature propagation module that captures temporal features from long-range videos while reducing memory usage.
Although exemplars can provide richer semantic information, exemplar-based approaches tend to lose the correlation between frames and exemplars when videos are lengthy or the motion is extensive, resulting in color artifacts.
To achieve stable and precise color control, SVCNet~\cite{zhao2023svcnet} proposes scribble-based video colorization, which includes CPNet for precise colorization and SSNet for temporal smoothing.
User-given color scribbles ensure precise region guidance and color control at the RGB level. 
However, scribbles can not express high-level semantics, such as \textit{"In winter..."} or \textit{"At sunset..."}. 
Therefore, relying solely on one modality—be it text, exemplars, or scribbles—makes it challenging to meet the demands of the task. 
To this end, we introduce a multi-modal solution that achieves high-quality and flexible video colorization by integrating multiple conditions.

%% file: sections/3_method.tex
\section{Method}\label{sec:method}

\subsection{Preliminaries}
Latent diffusion models (LDMs)~\cite{rombach2022high} are widely used in the image and video generation research community. 
LDMs learn to represent the distribution of images from large datasets by conducting the diffusion process and the denoising process in latent space.
The given image is first encoded into latent space by a VAE~\cite{kingma2013auto} encoder, after which Gaussian noise is added to the latent code. 
The noisy latent code is then fed into a UNet as input, which learns to reconstruct the clean image distribution at timestep $t$.
During this process, external conditions such as text are incorporated through the cross-attention mechanism to guide the denoising direction.
Finally, a VAE decoder decodes the clean latent code back to pixel space.
The overall training loss can be formulated as follows:
\begin{equation}
    \mathcal{L}_{\rm LDM} = \mathbb{E}_{\mathcal{E}(\mathbf{x}),\mathbf{y},\epsilon\sim\mathcal{N}(0, 1),t}\left[\lVert\epsilon - \epsilon_{\theta}(\mathbf{z}_t, t, \tau_{\theta}(\mathbf{y}))\rVert_2^2\right],
\end{equation}
where $\mathcal{E}(\cdot)$ is the VAE encoder, and $\mathbf{x}$, $\mathbf{y}$, $\tau_{\theta}(\cdot)$ and $\mathbf{z}_t$ represent the image, the text, the CLIP text encoder, and the noisy latent code at each timestep $t$, respectively.

\subsection{Network Design}
Given a grayscale video \(\mathbf{I}_{\rm g}^{1:N}=[I_{\rm g}^1, \cdots, I_{\rm g}^N]\), our goal is to synthesize a color video \(\mathbf{I}_{\rm c}^{1:N}=[I_{\rm c}^1, \cdots, I_{\rm c}^N]\), guided by the provided conditions such as hints, text, images, or a combination of these clues.

\begin{figure*}[!t]
    \centering
    \vspace{-6mm}
    \includegraphics[width=0.96\linewidth]
    {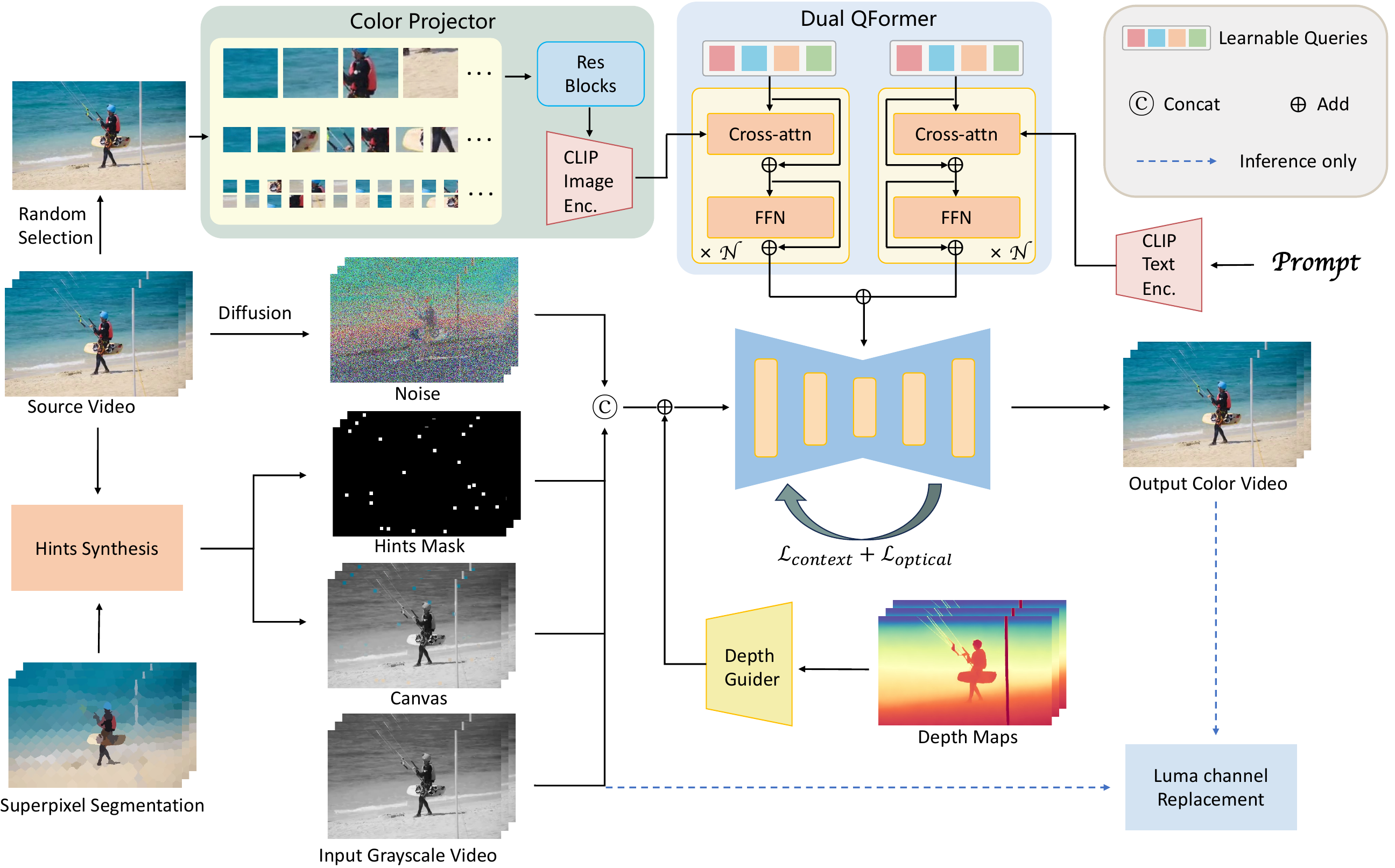}
    \caption{\textbf{Overall pipeline.} We omit the depiction of the VAE encoder and decoder for simplicity. Given a source color video $\mathbf{I}_{\rm gt}^{1:N}$, we first randomly select one frame as the exemplar image and feed it into the Color Projector, where the exemplar is divided into three groups of patches and then passed through the ResBlocks and CLIP image encoder to obtain the color features. The color features are sent to the Dual Qformer along with the encoded prompt, and the calculated features are injected into the UNet through cross attention. For hints injection, we leverage superpixel techniques to synthesize the hints mask $\mathbf{M}^{1:N}$ and the canvas $\mathbf{I}_{\rm canvas}$, which are concatenated with Gaussian noise and the grayscale video $\mathbf{I}_{\rm g}^{1:N}$ to serve as the input for the UNet. Additionally, we design a lightweight Depth Guider to enhance spatial-temporal consistency. During inference, we conduct luma channel replacement between the grayscale video and the output video to alleviate flickering artifacts caused by the video VAE.}
    \label{fig:pipeline}
\end{figure*}

As shown in~\cref{fig:pipeline}, we first convert the source color video \(\mathbf{I}_{\rm gt}^{1:N}=[I_{\rm gt}^1, \cdots, I_{\rm gt}^N]\) to obtain the grayscale video \(\mathbf{I}_{\rm g}\), then we randomly select one frame from the source video as an exemplar and input it into the Color Projector. 
The Color Projector output, along with the encoded prompt, is sent to the Dual QFormer.
Meanwhile, we calculate the superpixel segmentation of the source video to synthesize hints and utilize depth maps to enhance spatial-temporal consistency. 
During inference, we replace the luma channel of the output video with that of the grayscale video to eliminate flickering artifacts caused by the video VAE.

\noindent\textbf{Hints injection.}
Hints offer local guidance, enabling interactive control over colorization details.
Users only need to mark one or several hints in one frame of the grayscale video and specify the colors, and the entire video will be colorized based on the provided hints.

To achieve this, we first perform superpixel segmentation on the color video using the simple linear iterative clustering (SLIC)~\cite{khorasgani2022slic} algorithm, representing each superpixel by its mean color (denoted as \(\mathbf{I}_{\rm sp}\) $\in\mathbb{R}^{3 \times F \times H\times W}$). This approach retains color information while reducing fine-grained structural detail, resulting in a more uniform color distribution when sampling hints.
After that, we randomly select \textit{K} points on the first frame of the video and use CoTracker~\cite{karaev2023cotracker} to track these points in the video, thereby obtaining their trajectories.
For each point on the trajectory, we expand it into a cell centered at that point with a side length of $d$.
By integrating all cells in each frame, we get a mask sequence \(\mathbf{M}^{1:N} = [M^1, \cdots, M^k, \cdots, M^N]\), where \(M^k\) $\in\mathbb{R}^{H\times W}$, and \(M^k[i,j]=1\) denotes that a hint exists at the coordinate \((i,j)\) of the $k$-th frame; otherwise, \(M^k[i,j]=0\). 
We then compute the canvas as \(\mathbf{I}_{\rm canvas} = \mathbf{M} \times \mathbf{I}_{\rm sp} + (1-\mathbf{M}) \times \mathbf{I}_{\rm g} \).
Finally, \(\mathbf{I}_{\rm canvas}\) is encoded into latent space by a VAE encoder, and \(\mathbf{M}\) is downsampled and sliced with a stride of 4 to align with the shape of the latent code.
Their concatenation serves as the injection of hints of information.

\noindent\textbf{Color Projector.}
During training, a randomly selected frame from the video as an exemplar inherently contains structural information similar to the whole video.
However, we aim to allow the color distribution of an arbitrary image to be used as a reference, independent of structural similarity. 
To address this coupling of structure and color distribution in the training data, we design a Color Projector that effectively extracts the color features of the exemplar while weakening its structural information. 
We begin by randomly selecting a frame from the video and then dividing it into smaller patches at various scales, specifically at resolutions of $\frac{1}{4}$, $\frac{1}{8}$, and $\frac{1}{16}$.
After that, the multi-scale patches are sent to the Color Projector, where their patch embeddings are extracted using ResBlocks and then concatenated into a sequence.
To further enhance the color awareness between patch embeddings, we input this sequence into the CLIP~\cite{radford2021learning} image encoder and use the output as color features.

\begin{figure}[!t]
   
    \includegraphics[width=1.0\linewidth]
    {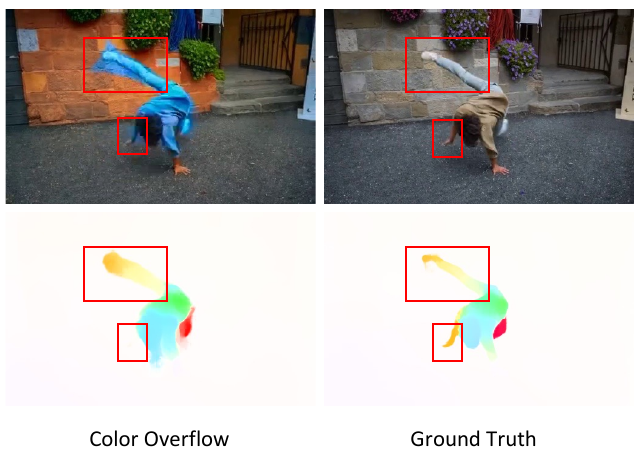}
    \vspace{-6mm}
    \caption{Color overflow caused by large motion results in optical flow estimation errors.}
    \label{fig:optical}
    \vspace{-6mm}
\end{figure}

\noindent\textbf{Dual Qformer.}
To bridge the natural gap between text and color feature modalities, we employ a Dual Qformer to map both the text and color representations into a shared feature space, facilitating their adaptation to SVD~\cite{blattmann2023stable}.
Specifically, text is first encoded into text embeddings by the CLIP text encoder. 
Subsequently, these text embeddings and color features are separately fed into the Qformer, which consists of a learnable query, \textit{N} stacked feed-forward network (FFN) layers, and cross-attention layers, allowing for effective cross-modal representation learning through the cross-attention mechanism.
After obtaining the learned queries \(\mathbf{l}_{\rm text}\) and \(\mathbf{l}_{\rm color}\), we apply two scaling factors, $\lambda_1$ and $\lambda_2$, and fuse the queries into \(\mathbf{l}_{\rm fuse} = \lambda_1 \times \mathbf{l}_{\rm text} + \lambda_2 \times \mathbf{l}_{\rm color}\), which is then fed into the cross-attention layers of SVD.


\noindent\textbf{Depth Guider.}
%
To reduce color overflow and improve spatial consistency in colorized results, we design a lightweight Depth Guider composed of several convolutional layers. 
We use Depth Anything V2~\cite{yang2024depth} to estimate depth information, which is then passed through the Depth Guider. 
The depth features are added to the latent code as input for the denoising UNet.

\begin{figure}[!b]
    \vspace{-4mm}
    \includegraphics[width=1.0\linewidth]
    {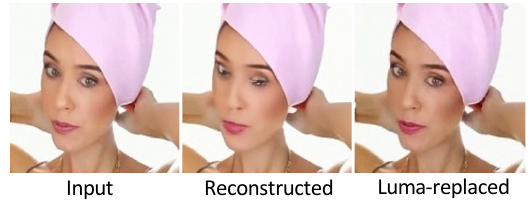}
    \vspace{-6mm}
    \caption{The reconstruction results of the video VAE exhibit flickering artifacts in high-frequency areas. Replacing the luma channel in $Lab$ color space can significantly improve the visual quality.}
    \label{fig:replaced}
    \vspace{-6mm}
\end{figure}
\begin{figure*}[!h]
     \vspace{-6mm}
    \includegraphics[width=1.0\linewidth]
    {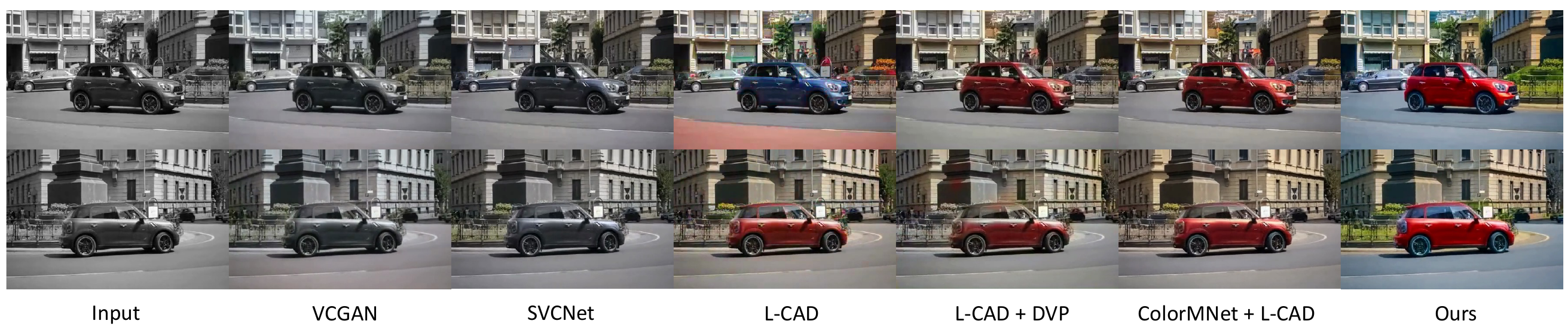}
    \vspace{-6mm}
    \caption{\textbf{Comparison results for automatic video colorization.} VCGAN and SVCNet exhibit severe grayish issues. L-CAD suffers from flickering artifacts; even though it is post-processed by DVP, color bleeding still persists. ColorMNet heavily relies on the colored exemplar frame, and error accumulation occurs, as we can see the top of the car turning black. In contrast, our model can generate temporal-coherent, vivid color videos.}
    \vspace{-8mm}
    \label{fig:auto}
\end{figure*}
\begin{figure}[!t]
    \vspace{3mm}
    \includegraphics[width=0.48\textwidth]
    {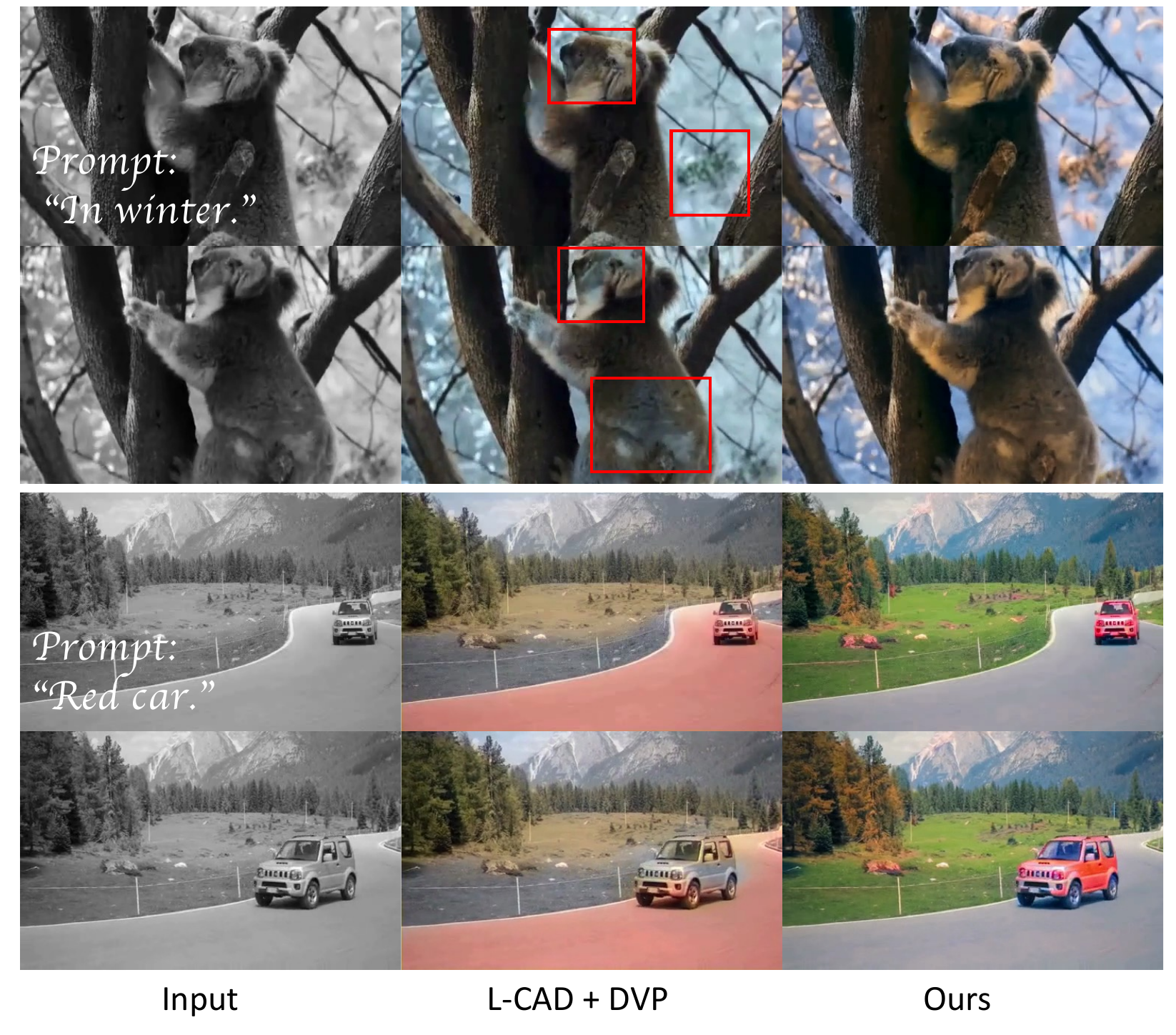}
     \vspace{-6mm}
    \caption{\textbf{Comparison for text-based video colorization.} L-CAD+DVP exhibits color bleeding and temporal incoherence. In contrast, our method can generate vivid and natural results that align with given prompts.}
    \label{fig:text}
      \vspace{-6mm}
\end{figure}

\subsection{Training Strategy}
To achieve a more efficient training process, we adopt a two-stage training strategy. 
In the first stage, we train the model on image datasets to ensure that SVD acquires the prior knowledge that is necessary for colorization. 
In the second stage, we fine-tune the model on video datasets to enhance temporal consistency.

\begin{figure}[!t]
  \vspace{3mm}
    \includegraphics[width=0.48\textwidth]
    {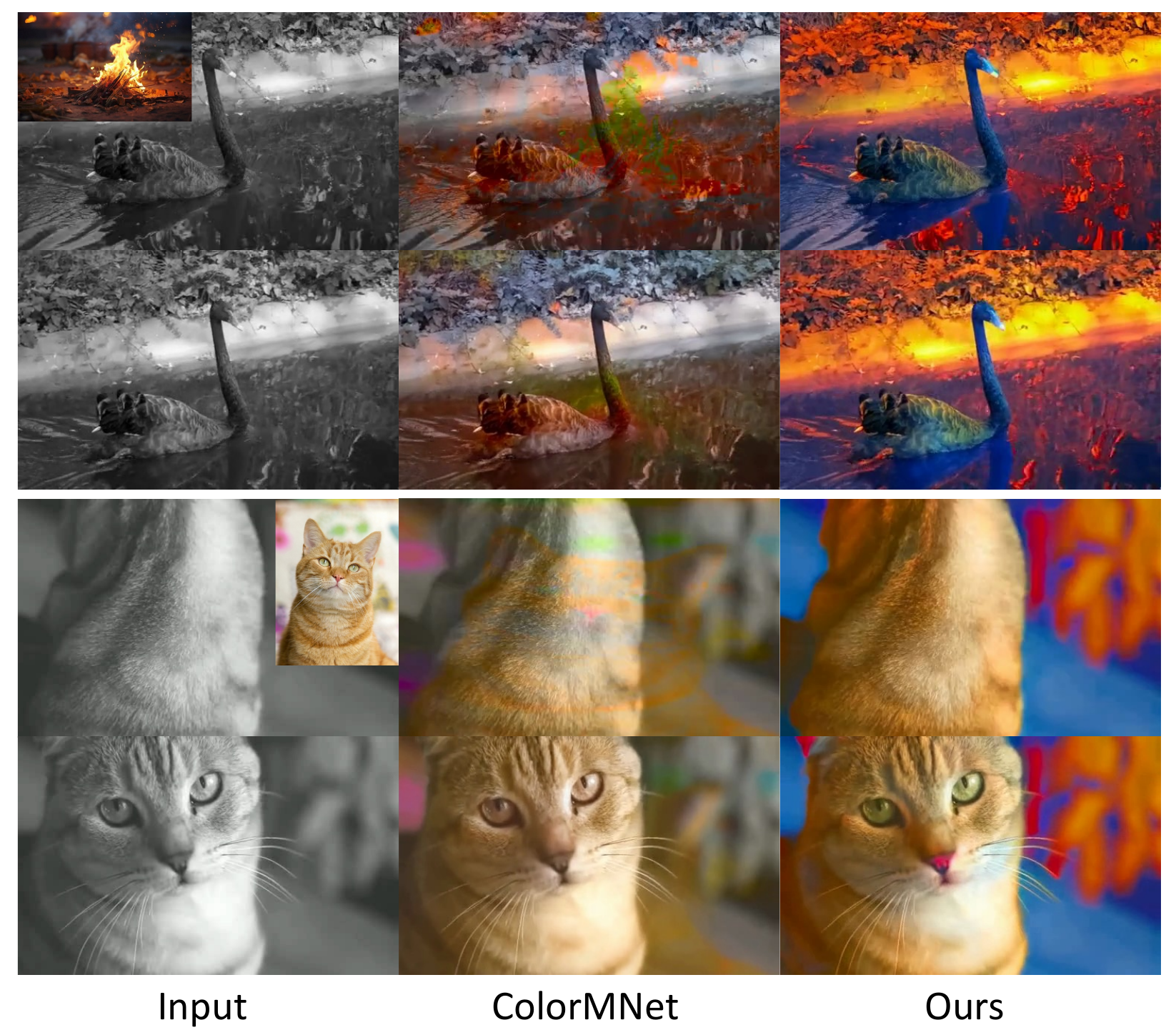}
    \vspace{-6mm}
    \caption{\textbf{Comparison for exemplar-based video colorization.} ColorMNet fails to produce exemplar-aligned videos. Our method is capable of capturing color distribution and semantic correspondence from exemplars.}
    \label{fig:exemplar}
    \vspace{-6mm}
\end{figure}

\noindent\textbf{Image stage.}
In the image training stage, we use the input image itself as the exemplar image and freeze the temporal modules of SVD to retain the knowledge of modeling the temporal dimension. 
Following Zhang \etal~\cite{zhang2019deep} and Liang \etal~\cite{liang2024control}, we employ contextual loss to further enhance the consistency of color distribution between the output image and the exemplar image.
Specifically, we have
\begin{equation}
\begin{aligned}
{d}^l(i,j)&=\mathtt{sim}_{\rm cos}(\phi^{l}_{I_{\rm gt}}(i),\phi^{l}_{I_{\rm c}}(j)),\\
A^l(i,j)&=\mathtt{softmax}_{j}(1-\tilde{d}^l(i,j)/h),\\ 
\mathcal{L}_{\rm context}&=\sum_{l} w_l[-\mathtt{log}(\frac{1}{N_l}\sum_i \mathtt{max}_j (A^l(i,j))],
\end{aligned}
\end{equation}
where $\phi^l$ represent the feature maps extracted at the
$relul\_2$ layer from the VGG19 network,
$\tilde{d}^l(i,j)$ denotes the normalized cosine similarity $d^l(i,j)$ of paired feature points, $A^l(i,j)$ is the pairwise affinities between features from the $l$-th layer from the VGG19 network, and $h$ and $w_l$ are hyperparameters.

\noindent\textbf{Video stage.}
After shifting SVD to the task of colorization, we fine-tune the temporal modules of SVD on the video dataset. 

\begin{figure}[!t]
    \includegraphics[width=0.48\textwidth]
    {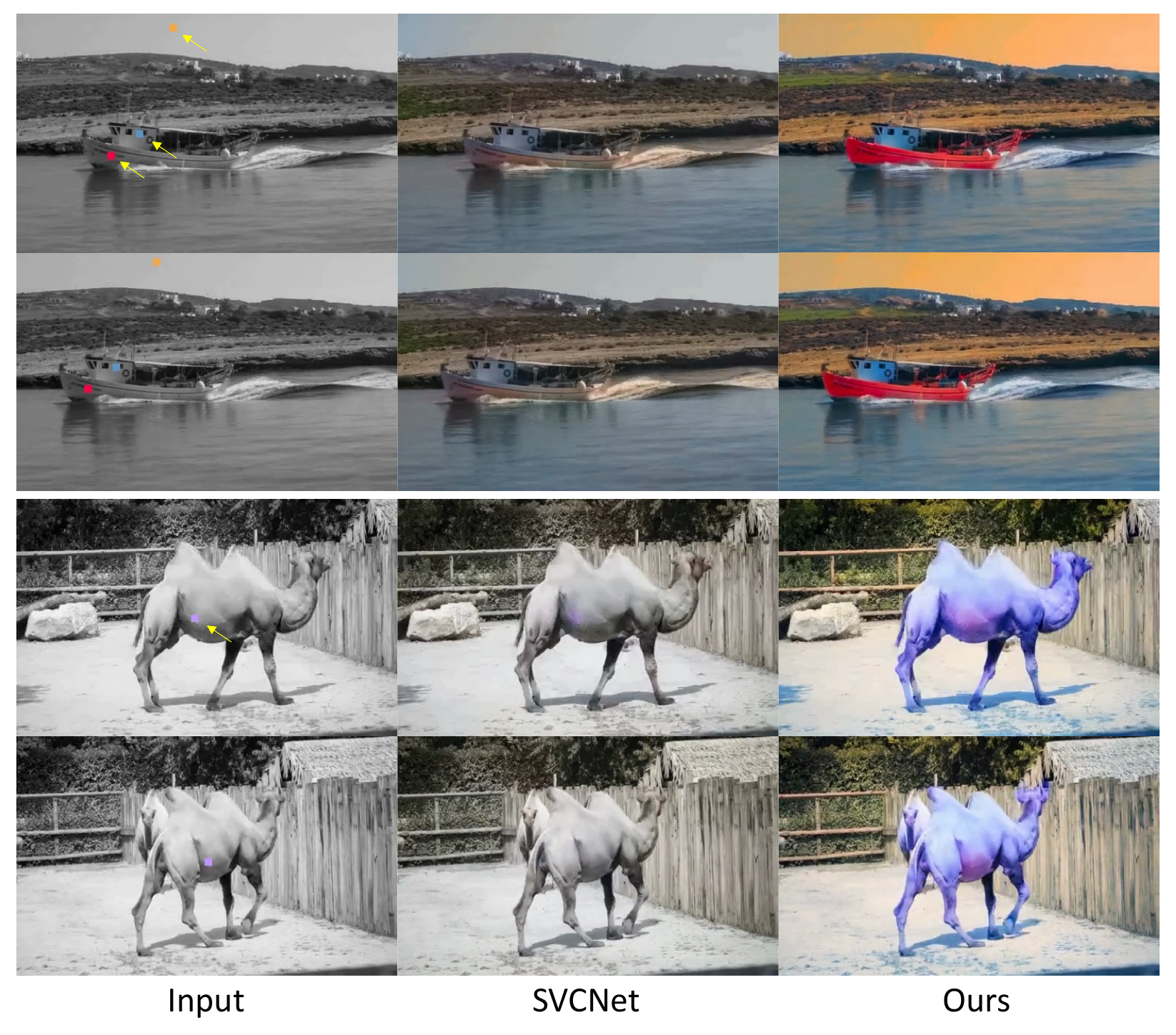}
    \vspace{-6mm}
    \caption{\textbf{Comparison for hints-based video colorization.} SVCNet suffers from grayish issues and fails to align with the given hints, while our method can synthesize diverse results that align with the provided hints.}
    \label{fig:hints}
    \vspace{-6mm}
\end{figure}

As shown in~\cref{fig:optical}, we observe that when color overflow occurs in the predicted video, even with the structural information preserved intact, the areas of color overflow still exhibit optical flow estimation errors.
To address this, we introduce an optical flow loss to mitigate color overflow caused by large motion.
Specifically, we select the $i$-th frame and the $(i+1)$-th frame from the source color video, and use GMFlow~\cite{xu2022gmflow} to compute the optical flow \(\mathbf{V}_{\rm gt}\) between the two frames.
We then calculate the optical flow \(\mathbf{V}_{\rm pre}\) from the predicted video using the same method. 
After that, we compute the MSE loss between the optical flows:
\begin{equation}
    \mathcal{L}_{optical} = \gamma \|\mathbf{V}_{\rm pre}-\mathbf{V}_{\rm gt}\|_2,
\end{equation}
where $\gamma$ is the weight hyperparameter.

\subsection{Luma Channel Replacement}


As shown in~\cref{fig:replaced}, we observe that video reconstruction with video VAEs often suffers from issues like pixel drifting in high-frequency areas, leading to noticeable flickering artifacts. 
Inspired techniques in image colorization~\cite{zabari2023diffusing,liang2024control,kang2023ddcolor}, we leverage the inherent stability of the grayscale video to maintain structural consistency in VAE reconstructions. 
Thus, we combine the luma channel of the grayscale video with the $ab$ channels of the predicted color video to produce more visually stable and pleasing results.

%% file: sections/4_exp.tex
\section{Experiments}\label{sec:exp}
\subsection{Implementation Details}
We collect an in-house image dataset containing 1.9M image-text pairs and choose OpenVid-1M~\cite{nan2024openvid} as our video dataset.
We utilize SVD-img2vid-xt~\cite{svd-xt} as the base model and CV-VAE~\cite{zhao2024cv} as the video VAE to generate longer videos with fewer computational resources.
During training, we resize the input data to $320 \times 512$ and use AdamW~\cite{loshchilov2017decoupled} as the optimizer, with a learning rate set to $1 \times 10^{-5}$.
We train our model on 4 Nvidia A800 GPUs. 
In the image stage, we set the batch size to 8 and the training steps to 130k. 
In the video stage, we set the batch size to 1, the video clip length to 61 frames, and the number of iterations to 50k.

For hints sampling, we randomly select \textit{K} points ranging from 0 to 150 and set the side length of the cells to be between 10 and 20.
During training, we set the scaling factors $ \lambda_1=1$ and $ \lambda_2=1$. For the contextual loss, we set $h=0.1$ and $w_l$ to $8, 4, 2$ for $l = 5, 4, 3$. 
For the optical flow loss, we set $\gamma=1$.

During inference, we optionally utilize SD-DINO~\cite{zhang2024tale} to map the exemplar image to the canvas in a frame-wise manner, thereby achieving strong semantic matching for the exemplar.


\begin{table*}[!t]
 \vspace{-6mm}
\caption{\textbf{Quantitative comparison for automatic colorization.} Our method is the most versatile while achieving state-of-the-art performance on most metrics.}
\vspace{-6mm}
  \begin{center}
  \resizebox{1\linewidth}{!}{%

  \begin{tabular}{c|cccccccccc}
    \toprule
      & Automatic & Text & Exemplar & Hints & SSIM$\uparrow$ & LPIPS$\downarrow$ & PSNR$\uparrow$ & Colorfulness$\uparrow$ & FVMD$\downarrow$ & Colorfulness / FVMD$\uparrow$\\
    \midrule
    VCGAN & $\checkmark$ & $\times$ & $\times$ & $\times$ & 
    0.9196 & 0.1912 & 22.5359 & 13.5129 & 645.6010 & 0.0209\\
    
    ColorMNet & $\times$ & $\times$ & $\checkmark$ & $\times$ & \textbf{0.9517} & 0.1956 & 22.2794 & 27.1427 & 632.0929 & 0.0429\\

    L-CAD & $\checkmark$ & $\checkmark$ & $\times$ & $\times$ &
    0.9184 & 0.2493 & 21.9502 & 34.5885 & 994.6452 & 0.0347\\

    L-CAD+DVP & $\checkmark$ & $\checkmark$ & $\times$ & $\times$ &
    0.9325 & 0.2354 & 22.5300 & 26.2798 & 779.8531 & 0.0336\\

    SVCNet & $\checkmark$ & $\times$ & $\times$ & $\checkmark$ &
    0.9294 & 0.1951 & 23.1570 & 14.2890 & 678.0615 & 0.0213\\

    Grayscle & \textbf{-} & \textbf{-} & \textbf{-} & \textbf{-} &
    0.9371 & 0.1962 & 23.0636 & 0.0000 & \textbf{596.1689} & 0.0000\\
      
    \midrule
    Ours & $\checkmark$ & $\checkmark$ & $\checkmark$ & $\checkmark$ & 0.9393 & \textbf{0.1908} & \textbf{23.2013} & \textbf{60.0881} & 662.9929 & \textbf{0.0906}\\
    \bottomrule
  \end{tabular}

  }
  \end{center}
  \label{tab:quant}
  \vspace{-5mm}
\end{table*}

\subsection{Qualitative Comparison}
We conduct qualitative comparison experiments with existing colorization methods, including automatic (VCGAN~\cite{zhao2022vcgan}), text-based (L-CAD~\cite{weng2024cad}), exemplar-based (ColorMNet~\cite{yang2025colormnet}), and hints-based (SVCNet~\cite{zhao2023svcnet}) solutions.

As shown in ~\cref{fig:auto}, regarding automatic colorization, both VCGAN~\cite{zhao2022vcgan} and SVCNet~\cite{zhao2023svcnet} suffer from severe grayish issues. 
Directly applying L-CAD~\cite{weng2024cad} frame by frame leads to significant temporal inconsistency and color flickering. 
After using Deep-Video-Prior (DVP)~\cite{lei2022deep} to post-process the results from L-CAD, although temporal consistency is improved, color overflow still persists. 
Since ColorMNet~\cite{yang2025colormnet} does not support automatic coloring, we first utilize L-CAD to obtain a colored exemplar frame, which is then fed to ColorMNet to enable automatic colorization capability.
The results, however, heavily depend on the quality of the exemplar frame and can lead to error accumulation. 
In contrast, our method can produce natural, color-rich videos without any assistance.

For text-based colorization, we compare our method with L-CAD + DVP. 
As shown in the first two rows of~\cref{fig:text}, for the prompt \textit{"In a snowy day, a koala is climbing the tree."}, L-CAD~\cite{weng2024cad} exhibits color bleeding issues, with abrupt color changes on the koala's face and unnatural color transitions in the thigh area due to the color bleeding effect created by the background colors. 
In the third and fourth rows, for the prompt \textit{"A red car."}, L-CAD generates unpleasant results with severe color bleeding. 
In contrast, our method generates text-aligned and temporally coherent results.

For exemplar-based colorization, our comparison results with ColorMNet are shown in ~\cref{fig:exemplar}. 
In the first two rows, ColorMNet~\cite{yang2025colormnet} fails to generate normal colorized results and suffers from significant artifacts. In the last two rows, ColorMNet exhibits grayish issues and color bleeding effects as the cat's eyes turn yellow. In contrast, our method can produce frames that align with the exemplar image and maintain good temporal consistency.

For hints-based colorization, we first select one or several hints in the first frame of the grayscale video.
The hints are then propagated throughout the entire video (using CPNet for SVCNet~\cite{zhao2023svcnet} and CoTracker~\cite{karaev2023cotracker} for our method).
As shown in the first two rows of ~\cref{fig:hints}, we select three hints on the boat and the sky; although SVCNet can generate natural color videos, it cannot respond to the given hints. 
In the last two rows, however, with only one hint as input, SVCNet exhibits grayish effects.
In contrast, our method can effectively respond to the provided hints and achieves diverse coloring of various regions with only a minimal number of hints.

\subsection{Quantitative Comparison}
To ensure a fair comparison, we conduct a quantitative evaluation of our method against existing methods on the DAVIS dataset~\cite{Perazzi2016} in an unconditional manner, using SSIM, LPIPS, PSNR, Colorfulness~\cite{hasler2003measuring}, and Fréchet video motion distance (FVMD)~\cite{liu2024fr} to assess video quality.
Additionally, we argue that colorfulness and temporal consistency should be evaluated as a unified metric for the task of video colorization, as this task not only requires temporal consistency but also demands diversity and richness in color. 
If Colorfulness and FVMD are used separately to assess video quality, it can lead to the following issues: 1) Each frame may exhibit very good visual quality, but the inter-frame relationships could be completely ignored; 
2) Grayish videos tend to achieve higher FVMD scores. 
As observed in our experiments, the FVMD metric for grayscale videos is the best, yet it does not satisfy the task of colorization. 
Therefore, we propose a new metric, denoted as Colorfulness / FVMD, which effectively takes into account both color and temporal relationships.
As shown in ~\cref{tab:quant}, our method is not only the most versatile but also achieves state-of-the-art performance across multiple metrics.
\begin{table}[!h]
\vspace{-2mm}
    \caption{\textbf{Quantitative comparison for text-based colorization.} Our method surpasses L-CAD in both CLIP score and Colorfulness.}
\vspace{-8mm}
  \begin{center}
  \resizebox{0.7\linewidth}{!}{%

  \begin{tabular}{c|cc}
    \toprule
      & Colorfulness$\uparrow$ & CLIP score$\uparrow$\\
    \midrule
       L-CAD & 43.0182 & 62.9091\\ 
        Ours & \textbf{62.1264} & \textbf{65.0381}\\ 
     \bottomrule
    \end{tabular}
  }
  \end{center}
   
  \label{tab:text_quant}
  \vspace{-6mm}
\end{table}

\begin{table}[!h]
\caption{\textbf{Quantitative comparison for exemplar-based colorization.} We outperform ColorMNet in both LPIPS and Colorfulness, demonstrating the robustness of our method.}
\vspace{-5mm}
  \begin{center}
  \resizebox{0.7\linewidth}{!}{%

  \begin{tabular}{c|cc}
    \toprule
      & LPIPS$\downarrow$ & Colorfulness$\uparrow$\\
    \midrule
      ColorMNet & 0.6437 & 38.9738\\ 
        Ours & \textbf{0.5693} & \textbf{65.9782}\\ 
     \bottomrule
    \end{tabular}
  }
  \end{center}
   
  \label{tab:exemplar_quant}
  \vspace{-5mm}
\end{table}

\begin{table}[!h]
\caption{\textbf{Quantitative comparison for hints-based colorization.} Our method achieves better performance in hints-guided colorization compared to SVCNet.}
\vspace{-5mm}
  \begin{center}
  \resizebox{0.7\linewidth}{!}{%

  \begin{tabular}{c|cc}
    \toprule
      & SSIM$\uparrow$ & Colorfulness$\uparrow$\\
    \midrule
       SVCNet & 0.9302 & 19.9826\\ 
        Ours & \textbf{0.9418} & \textbf{63.6238}\\ 
     \bottomrule
    \end{tabular}
  }
  \end{center}
   
  \label{tab:hints_quant}
  \vspace{-8mm}
\end{table}

\begin{figure*}[!t]
    \vspace{-6mm}
    \includegraphics[width=1\linewidth]
    {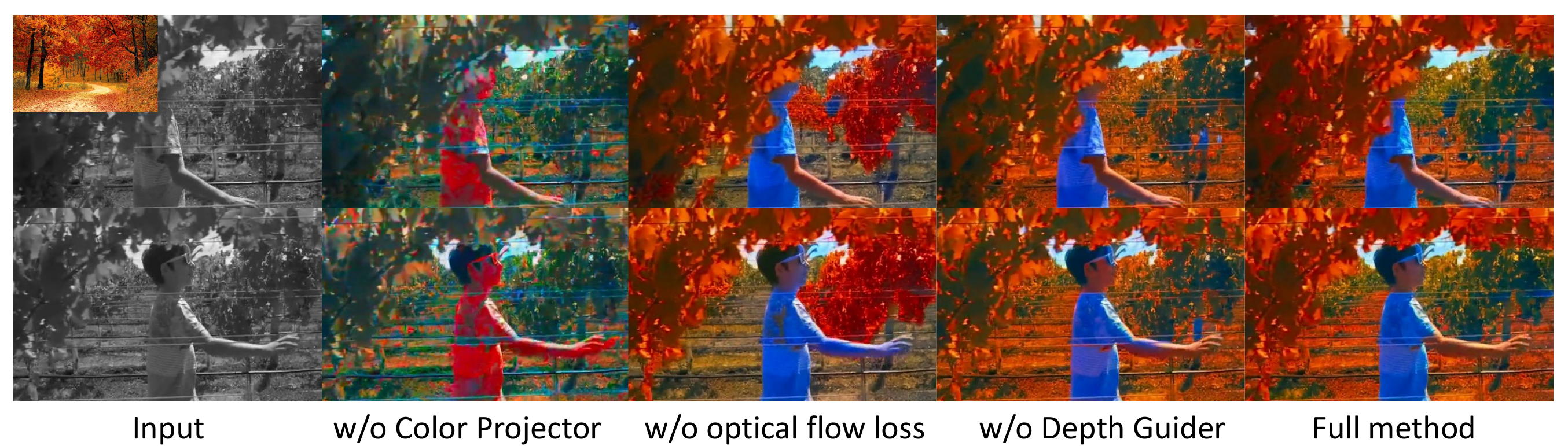}
    \caption{\textbf{Ablation studies.} Removing the Color Projector results in inaccurate color allocation. Without the optical flow loss, color overflow effects occur. Eliminating the Depth Guider leads to spatial-temporal inconsistency. In contrast, our full method achieves smooth and condition-aligned videos compared to other designs.}
    \vspace{-4mm}
    \label{fig:ablation}
\end{figure*}
For text-based comparison, we first leverage video captioning models to obtain descriptions of colored objects for the DAVIS\cite{Perazzi2016} validation set, then we employ CLIP score~\cite{hessel2021clipscore} to evaluate the coherence between the provided text and the generated videos. 
As demonstrated in ~\cref{tab:text_quant}, our method achieves a higher CLIP score and Colorfulness compared to L-CAD\cite{weng2024cad}. For exemplar-based methods, we collect a set of diverse images as exemplars, and we compute the LPIPS between the exemplar images and synthesized videos. As shown in ~\cref{tab:exemplar_quant}, our method surpasses ColorMNet\cite{yang2025colormnet} in both LPIPS and Colorfulness, demonstrating the robustness of our algorithm. For hints-based approaches, we compare our method with SVCNet\cite{zhao2023svcnet}. Specifically, we first conduct color augmentation for the source color videos to avoid automatic colorization priors; then we randomly sample hints from the augmented videos as input for SVCNet and our method. As illustrated in ~\cref{tab:hints_quant}, we outperform SVCNet in both SSIM and Colorfulness.


\subsection{Ablation Study}
To validate the effectiveness of our design, we conduct ablation studies on the components of the model.
As shown in ~\cref{fig:ablation}, given the prompt \textit{“A boy wearing a blue T-shirt.”} along with an exemplar image describing autumn maple leaves. 
Removing the Color Projector results in ineffective extraction of color information from the exemplar. 
Furthermore, omitting the optical flow loss leads to color overflow during significant motion, as seen with the arms turning blue. 
Without the Depth Guider, color bleeding also occurs.
Our complete method, however, can generate results that align with the input conditions and exhibit good temporal consistency.
We also examine the effectiveness of the Dual Qformer, please refer to the supplementary materials for more details on the ablation studies.

\subsection{User Study}
\begin{figure}[!t]
    \centering
    \vspace{-4mm}
    \includegraphics[width=0.8\linewidth]
    {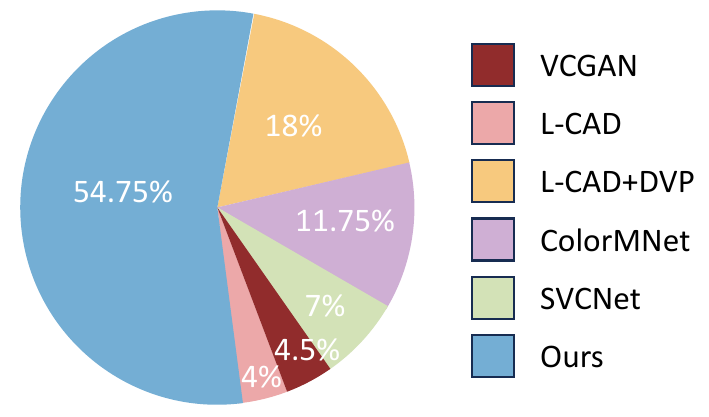}
    \caption{\textbf{User study results.} The vast majority of users prefer our model.}
    \label{fig:user}
    \vspace{-6mm}
\end{figure}

To further assess the quality of the generated results, we conduct a user study, as subjective evaluations from users are often a more reasonable standard in the field of colorization. 
Specifically, we prepare 20 questions that encompass several dimensions: color richness, temporal consistency, aesthetic preference, and the alignment between the given conditions (text, exemplars, and hints) and the synthesized videos.
In each question, participants are asked to choose the best result from various compared methods.
We receive 20 questionnaires, and the results are shown in ~\cref{fig:user}. 
Our method acquires the preference of the majority of users, followed by L-CAD+DVP and ColorMNet, while L-CAD and VCGAN are the least voted.

%% file: sections/5_conclusion.tex
\section{Conclusion}\label{sec:conclusion}
In this paper, we present VanGogh, a unified diffusion-based framework for multimodal video colorization. 
Comprehensive experiments and user studies demonstrate that our method is capable of generating visually appealing results that align with given conditions, thus significantly enhancing the flexibility and interactivity of the video colorization process.
We believe our methodology can serve as a valuable reference for researchers in the field of video colorization.

%% file: sections/6_ref.tex
{
\small
\bibliographystyle{ieeenat_fullname}
\bibliography{ref.bib}
}

%% file: sections/7_appendix_1.tex
\clearpage
\appendix
\renewcommand\thesection{\Alph{section}}
\renewcommand\thefigure{S\arabic{figure}}
\renewcommand\thetable{S\arabic{table}}
\renewcommand\theequation{S\arabic{equation}}
\setcounter{figure}{0}
\setcounter{table}{0}
\setcounter{equation}{0}
\setcounter{page}{1}
\maketitlesupplementary
In this supplementary material, we first describe the specific details of our implementation in~\cref{imple}, including the parameter settings of the network during training and the analysis of luma channel replacement. Then, in~\cref{ablation}, we present more detailed ablation experiments and an analysis of the Dual Qformer. Finally, in~\cref{vis}, we show more visualization results.


\section{Implementation Details}\label{imple}
\subsection{Training details}\label{training_details} 
We use SVD-img2vid-xt~\cite{svd-xt} to initialize the UNet. The original SVD~\cite{blattmann2023stable} has 4 input channels, but since we concatenated the hints mask, canvas, and grayscale video to the original latent code, the input channels are expanded to 13 (4 for the original latent code, 1 for the hints mask, 4 for the canvas, and 4 for the grayscale video). During training, we modify the input channels of the UNet $conv\_in$ layer with $in\_channel=13$ and randomly initialize the parameters. 

For the selection of control conditions, we set a random variable \( X \sim U(0,1) \). When \( X  \in [0, 0.3)\), we only use the image modality as the condition, and both the hints information and the text are set to null. Similarly, when \( X  \in [0.3, 0.5) \), we only use text as the condition, and when \( X \in [0.5, 0.8) \), we only keep the hints guidance.
For multi-condition control, when \( X \in [0.8, 0.83) \), we preserve the image and hints; when \( X \in [0.83, 0.86) \), we keep the text and hints; when \( X \in [0.86, 0.9) \), we maintain the image and text. When \( X \in [0.9, 0.95) \), all three modalities are retained, and finally, when \( X \in [0.95, 1] \), all three modalities are discarded.


\subsection{Luma channel replacement}\label{luma} 
The $Lab$ color space (often referred to as $CIELAB$ or simply $LAB$) is a color model that is designed to be a more perceptually uniform representation of colors than other color spaces like RGB or CMYK. The key components of the $Lab$ color space are: 

1) Luma: This represents the lightness of the image, ranging from 0 (black) to 100 (white). 

2) a: This axis represents the color's position between green and red. Negative values indicate green, and positive values indicate red.

3) b: This axis represents the position between blue and yellow. Negative values indicate blue, and positive values indicate yellow.

In the task of colorization, the luma channel of the grayscale input retains the structural information intact, which is why many image-based colorization methods~\cite{kang2023ddcolor,liang2024control,zabari2023diffusing} replace the luma channel of the colorized output with the luma channel of the grayscale input to enhance image quality. However, we find that this technique can also address the flickering artifacts caused by the VAE video VAE~\cite{zhao2024cv}, thus not only improving image quality but also enhancing temporal consistency. As shown in Fig. 4, we observe that the reconstruction results of the VAE show obvious artifacts. After replacing the luma channels, the quality improves significantly. Thus, we argue that the human eye is more sensitive to structural artifacts than to color artifacts.

\section{Ablation study}\label{ablation}


\begin{table*}[!t]
 \vspace{-6mm}
\caption{\textbf{Quantitative comparison for ablation studies.} The full method outperforms other designs across all the metrics.}
\vspace{-6mm}
  \begin{center}
  \resizebox{1\linewidth}{!}{%

  \begin{tabular}{c|cccccc}
    \toprule
      & SSIM$\uparrow$ & LPIPS$\downarrow$ & PSNR$\uparrow$ & Colorfulness$\uparrow$ & FVMD$\downarrow$ & Colorfulness / FVMD$\uparrow$\\
    \midrule
    w/o contextual loss & 0.9154 & 0.1990 & 22.4264 & 56.2335 & 674.9781 & 0.0833\\
    
    w/o optical flow loss & 0.9023 & 0.2271 & 21.2794 & 57.8907 & 721.2509 & 0.0802\\

    w/o Color Projector & 0.9278 & 0.2053 & 22.8653 & 53.9126 & 671.0116 & 0.0803\\

    w/o Depth Guider & 0.9124 & 0.2123 & 21.1537 & 58.3792 & 698.4211 & 0.0835\\

    w/o Dual QFormer & 0.9302 & 0.2019 & 22.5060 & 55.7230 & 683.8873 & 0.0814\\
      
    \midrule
    Full method & \textbf{0.9393} & \textbf{0.1908} & \textbf{23.2013} & \textbf{60.0881} & \textbf{662.9929} & \textbf{0.0906}\\
    \bottomrule
  \end{tabular}

  }
  \end{center}
  \label{tab:quant}
  \vspace{-5mm}
\end{table*}

\begin{figure*}[!t]
    \centering
    \includegraphics[width=1\linewidth]{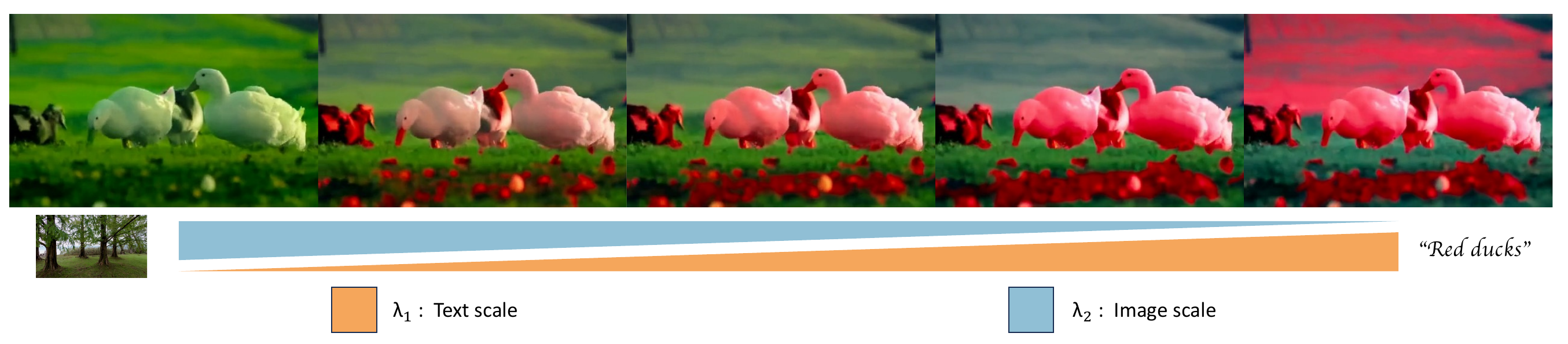}
    \vspace{-6mm}
    \caption{%
        \textbf{Scaling factor control.} By modifying the text scale and image scale, our method can generate results with varying degrees of alignment between text and image. When the text scale increases, the duck turns red, while when the image scale increases, the background turns green.
    }
    \label{fig:scale}
\end{figure*}
\subsection{Quantitative results}\label{quant}
We conduct ablation studies on the components of the mode across the following metrics: SSIM, LPIPS, PSNR, Colorfulness~\cite{hasler2003measuring}, FVMD~\cite{liu2024fr} and Colorfulness / FVMD. Note that we do not adopt FVD~\cite{Unterthiner2018TowardsAG} because FVD can only measure whether the appearance of the generated video is consistent with the reference video. However, video colorization is inherently an ill-posed problem, and evaluating the quality of the generated results based on whether the appearance aligns with the ground truth is biased. Furthermore, our experiments reveal that after post-processing with DVP~\cite{lei2022deep}, the results of L-CAD~\cite{weng2024cad}+DVP perform worse on the FVD metric than L-CAD, which colorizes each frame individually. This is clearly unreasonable.

We evaluate our designs on the DAVIS validation dataset~\cite{Perazzi2016} in an automatic manner. As shown in~\cref{tab:quant}, the full method outperforms other designs in all metrics.

\begin{figure*}[!t]
    \centering
    \vspace{-8mm}
    \includegraphics[width=0.9\linewidth]{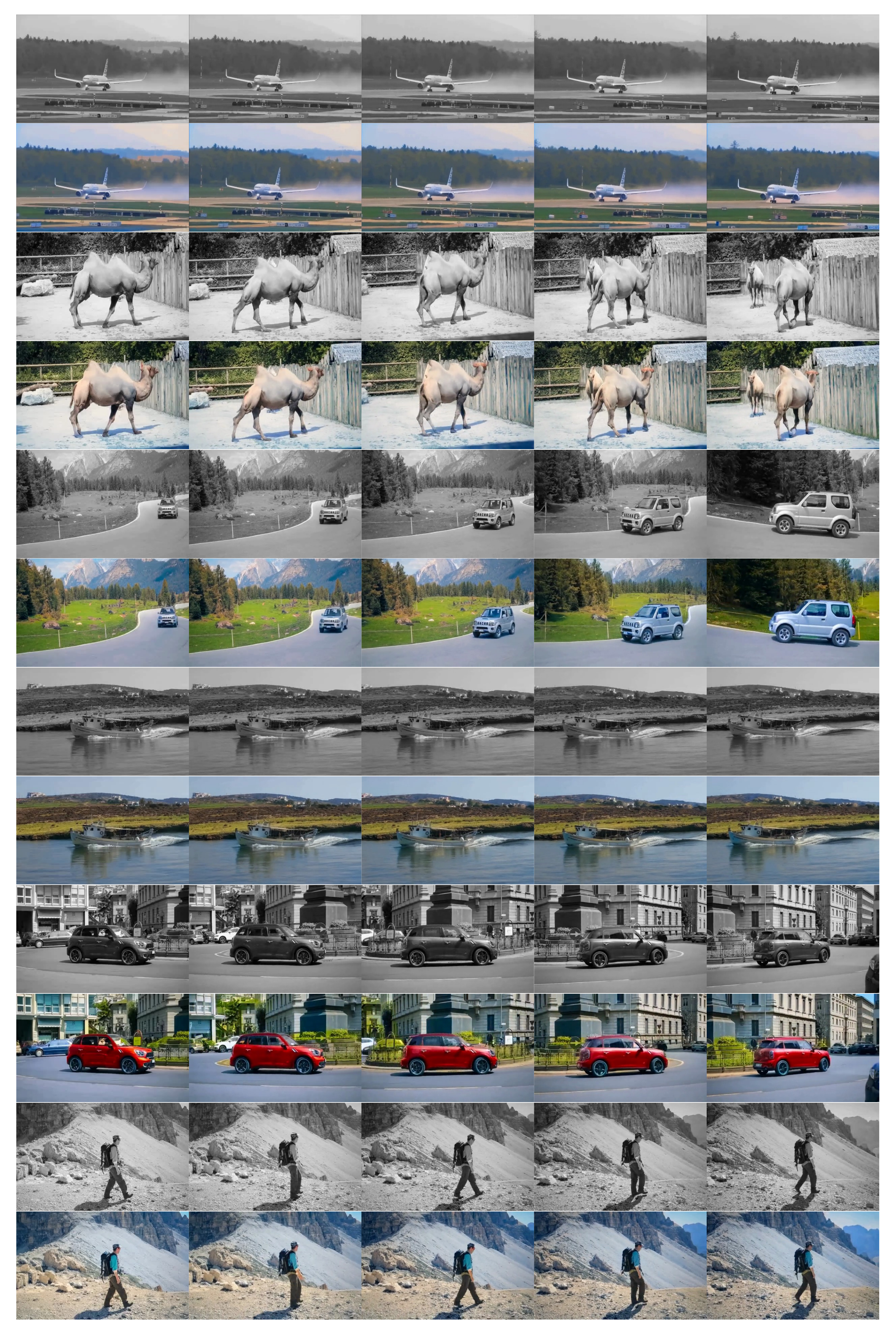}
    \vspace{-4mm}
    \caption{%
        \textbf{Automatic colorization results.} Each pair of two rows contains a gray scale video and a colorized video. Our method is capable of generating natural, vivid color videos.
    }
    \label{fig:automatic1}
\end{figure*}

\subsection{Analysis on the Dual QFormer}\label{dqormer}
We propose the Dual QFormer to fill the gap between image modality and text modality and force the adaption of SVD~\cite{blattmann2023stable} to the task of colorization. The learnable queries extract color semantics from both the text and the image effectively, and fuse them into a shared feature space, enhancing the alignment between the colorization results and the given conditions. For text-based colorization, we leverage CLIP score\cite{hessel2021clipscore} to assess the alignment between the colorization videos and the provided text. As shown in~\cref{tab:text_ab_quant}, Dual QFormer offers better perceptual capabilities in text description. For exemplar-based colorization, we compare the full method with both the Dual QFormer-removed version and the Color Projector-removed version. As shown in~\cref{tab:exemplar_ab_quant}, removing either the Dual QFormer or the Color Projector results in the insufficient extraction of the color semantics of the exemplar, thereby compromising the alignment between the coloring results and the exemplar images.

\begin{table}[!h]
\vspace{-2mm}
    \caption{\textbf{Quantitative comparison for text-based ablation.} The Dual QFormer enhances performance in both CLIP score and Colorfulness.}
\vspace{-6mm}
  \begin{center}
  \resizebox{0.7\linewidth}{!}{%

  \begin{tabular}{c|cc}
    \toprule
      & Colorfulness$\uparrow$ & CLIP score$\uparrow$\\
    \midrule
       w/o Dual QFormer & 60.3253 & 61.0973\\ 
        Full method & \textbf{62.1264} & \textbf{65.0381}\\ 
     \bottomrule
    \end{tabular}
  }
  \end{center}
   
  \label{tab:text_ab_quant}
  \vspace{-6mm}
\end{table}

\begin{table}[!h]
\vspace{-2mm}
    \caption{\textbf{Quantitative comparison for exemplar-based ablation.} The Dual QFormer and Color Projector both enhance performance in LPIPS score and Colorfulness.}
\vspace{-6mm}
  \begin{center}
  \resizebox{0.7\linewidth}{!}{%

  \begin{tabular}{c|cc}
    \toprule
      & LPIPS$\downarrow$ & Colorfulness$\uparrow$\\
    \midrule
        w/o Color Projector & 0.5912 & 52.9806\\ 
       w/o Dual QFormer & 0.6023 & 50.4281\\ 
        Full method & \textbf{0.5693} & \textbf{65.9782}\\ 
     \bottomrule
    \end{tabular}
  }
  \end{center}
   
  \label{tab:exemplar_ab_quant}
  \vspace{-6mm}
\end{table}

As depicted in Sec. 3.2, we set two scaling factors, $\lambda_1$ and $\lambda_2$, to fuse the extracted color features and text features into a shared feature space. With this design, users can control the intensity of the given exemplar image and text, thereby enhancing the flexibility of multi-condition colorization. As shown in~\cref{fig:scale}, given the prompt \textit{"Red ducks"} and the exemplar image, modifying the text scale ($\lambda_1$) and the image scale ($\lambda_2$) in different ratios yields diverse colorized results, demonstrating that Dual QFormer offers a unified feature space and thus improves the interactivity and flexibility.

\section{More Qualitative Results}\label{vis}

\subsection{Automatic colorization results}\label{auto}
As shown in~\cref{fig:automatic1}, we present more automatic colorization results. Our method can directly generate realistic, natural, and richly colored videos without any conditions.

\subsection{Single-condition colorization results}\label{single}
\begin{figure*}[!t]
    \centering
    \vspace{-8mm}
    \includegraphics[width=0.9\linewidth]{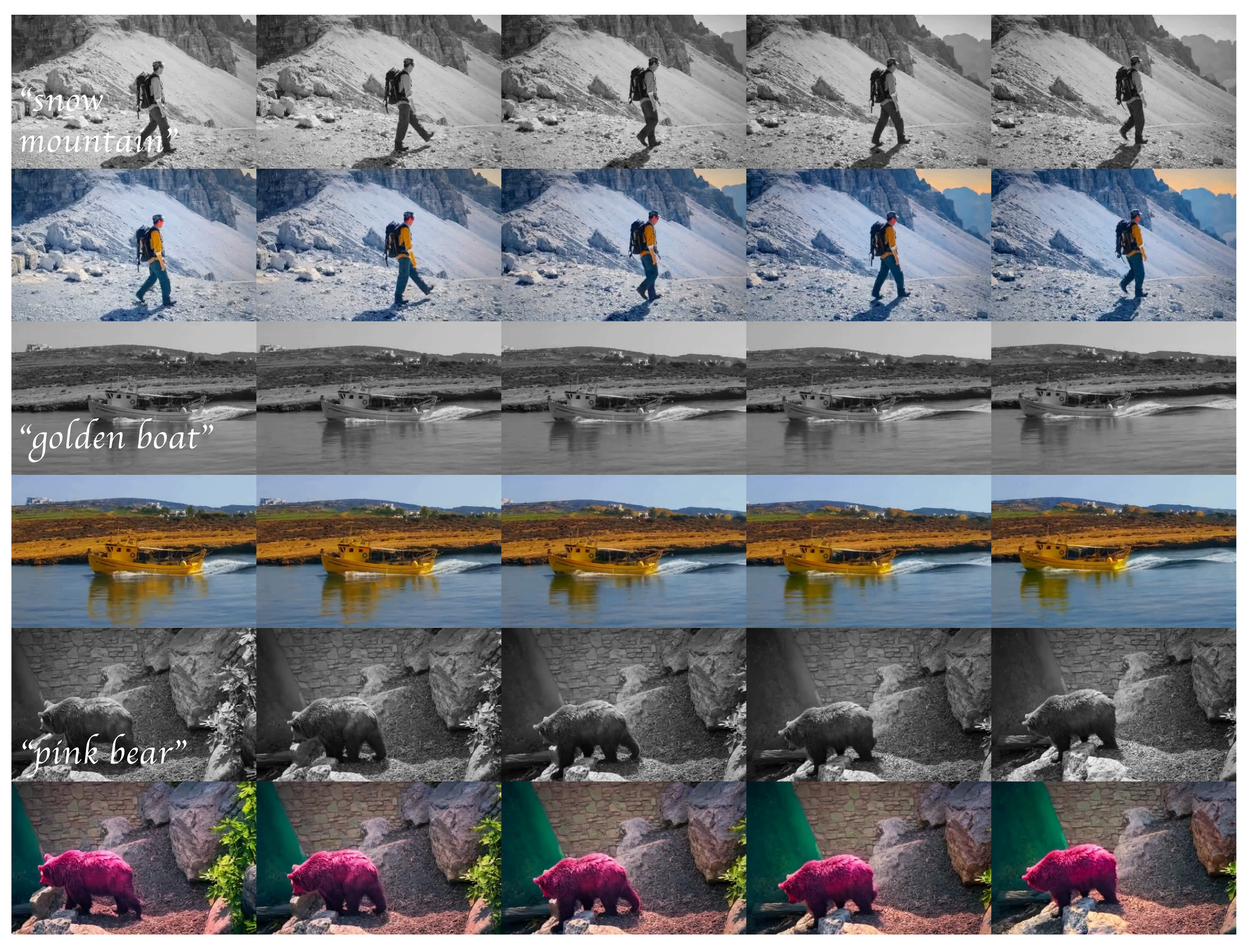}
    \vspace{-4mm}
    \caption{%
        \textbf{Text-based colorization results.} In every two rows, the first row contains a grayscale video and a prompt, while the second row shows the colorized result. Our method can generate color videos that align with the text.
    }
    \label{fig:text1}
\end{figure*}

\begin{figure*}[!t]
    \centering
    \vspace{-8mm}
    \includegraphics[width=0.9\linewidth]{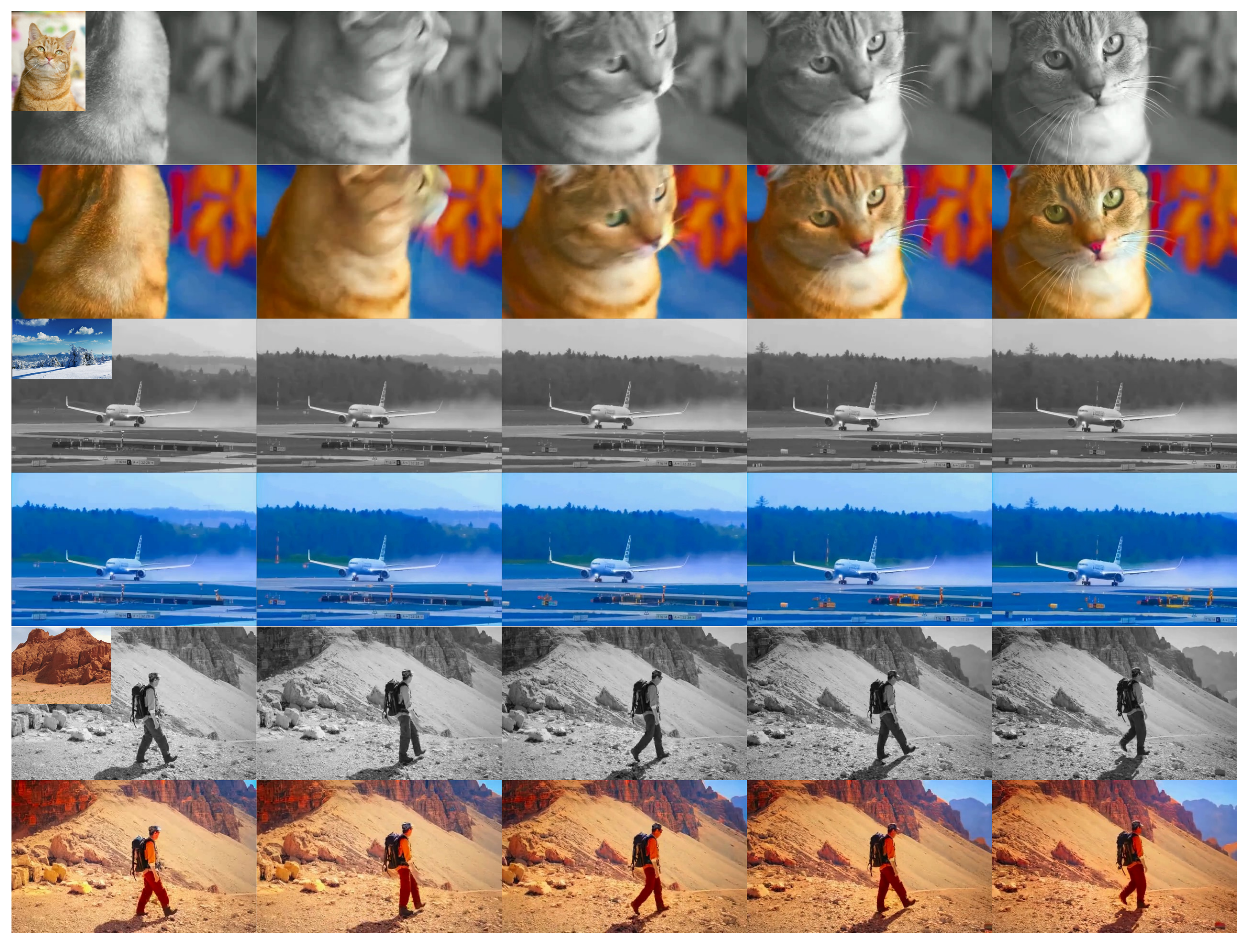}
    \vspace{-4mm}
    \caption{%
         \textbf{Exemplar-based colorization results.} In every two rows, the first row contains a grayscale video and an exemplar image, while the second row shows the colorized result. Our method can generate color videos that align with the exemplars.
    }
    \label{fig:exemplar1}
\end{figure*}

\begin{figure*}[!t]
    \centering
    \vspace{-8mm}
    \includegraphics[width=0.9\linewidth]{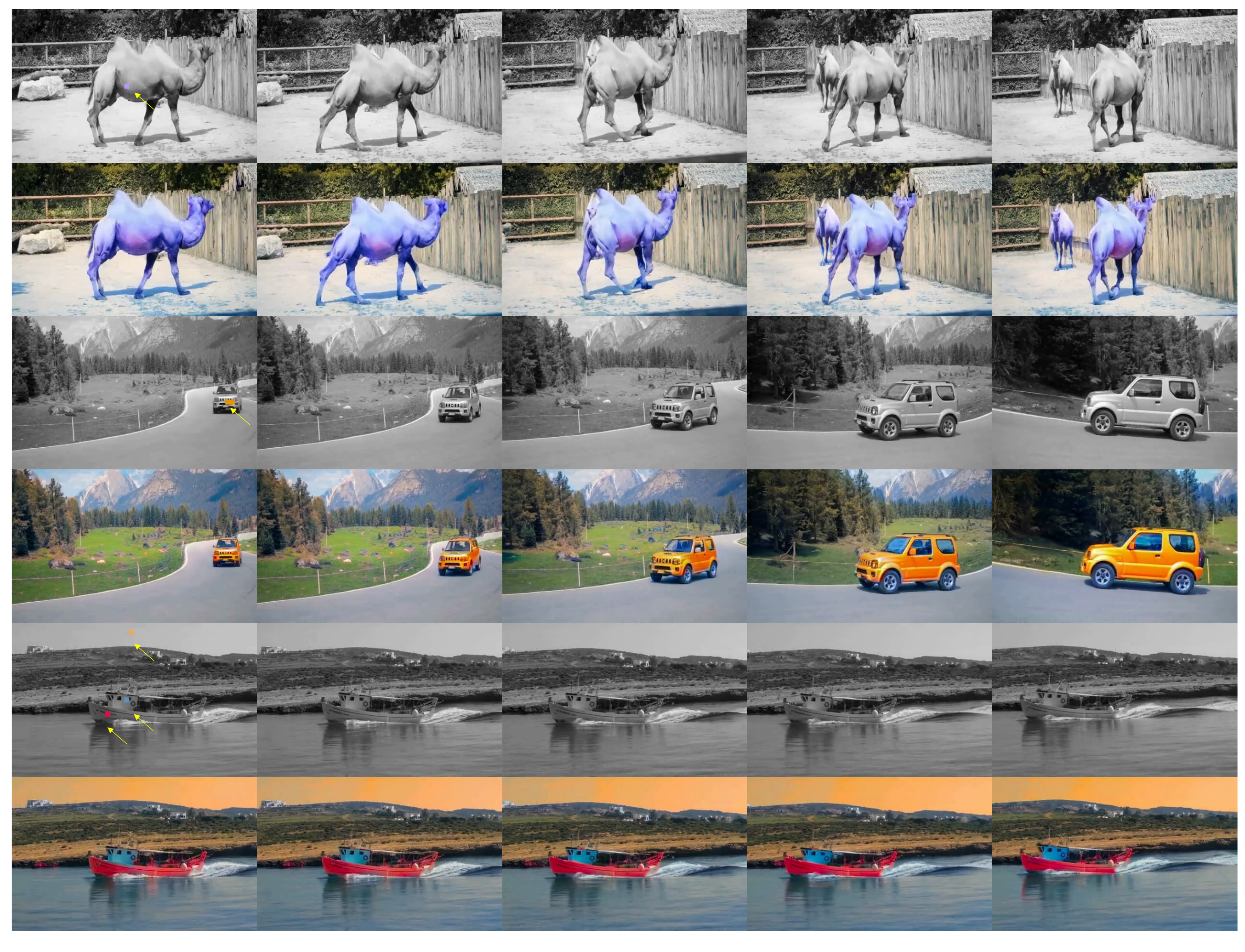}
    \vspace{-4mm}
    \caption{%
         \textbf{Hints-based colorization results.} In every two rows, the first row contains a grayscale video along with the provided hints (see yellow arrow(s)), while the second row shows the colorized result. Our method can generate color videos that align with the hints.
    }
    \label{fig:hints1}
\end{figure*}

We also include the colorization videos under single-condition guidance. As shown in~\cref{fig:text1}, ~\cref{fig:exemplar1}, and ~\cref{fig:hints1}, we display the results for text, exemplar, and hints, respectively. Our method can synthesis condition-aligned color videos.

\subsection{Multi-condition colorization results}\label{multi}
\begin{figure*}[!t]
    \centering
    \vspace{-8mm}
    \includegraphics[width=0.9\linewidth]{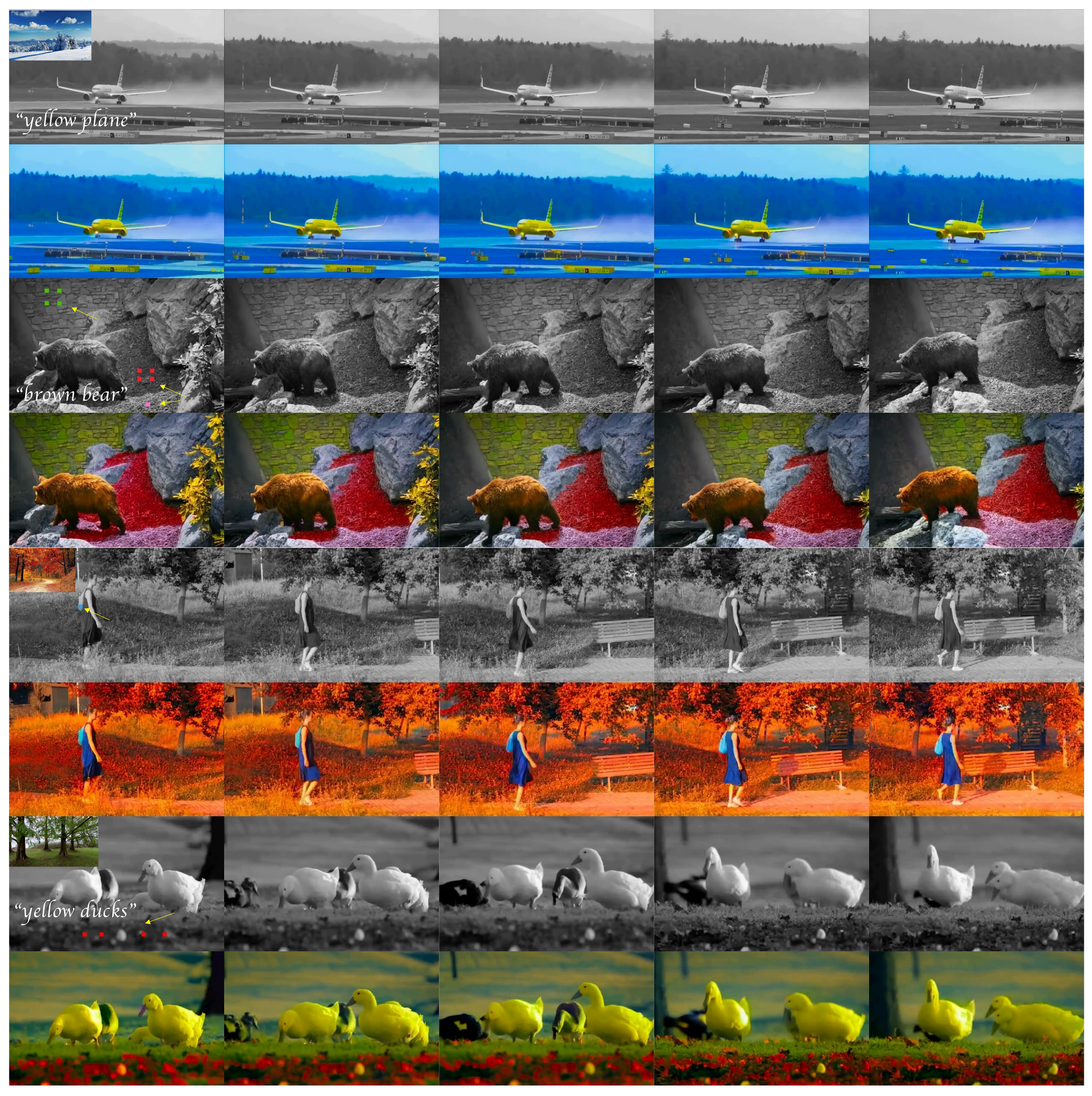}
    \vspace{-4mm}
    \caption{%
        \textbf{Multi-condition colorization results.} The first two rows display the colorization results of text + exemplar. The third and fourth rows show the results of text + hints, the fifth and sixth rows display the results of exemplar + hints, and the last two rows present the colorization videos of text + exemplar + hints.
    }
    \label{fig:multi}
\end{figure*}

Our model can accept multiple conditions to achieve global-local joint control, as shown in~\cref{fig:multi}. Given various conditions, our method can generate diverse results that align with the given conditions, significantly improving the visual quality, interactivity, and flexibility of video colorization.

